\documentclass[lettersize,journal]{IEEEtran}
\usepackage{amsmath,amsfonts}
\usepackage{array}
\usepackage{booktabs}
\usepackage{gensymb}
\usepackage{textcomp}
\usepackage{stfloats}
\usepackage{url}
\usepackage{verbatim}
\usepackage{graphicx}
\usepackage{cite}
\usepackage{amsmath}
\usepackage{amssymb}
\usepackage{breqn}
\usepackage{bm}
\usepackage{pgf}
\usepackage{tabularx}
\usepackage{textcomp}
\usepackage{color}
\usepackage{subcaption}
\PassOptionsToPackage{hyphens}{url}
\usepackage[hidelinks]{hyperref}
\usepackage{tikz}
\usetikzlibrary{decorations.pathreplacing,calc,shapes.geometric,matrix,trees,positioning}
\usepackage[ruled,linesnumbered]{algorithm2e}
\SetAlgoLined
\makeatletter
\newcommand{\nosemic}{\renewcommand{\@endalgocfline}{\relax}}
\newcommand{\dosemic}{\renewcommand{\@endalgocfline}{\algocf@endline}}
\let\oldnl\nl
\newcommand{\nonl}{\renewcommand{\nl}{\let\nl\oldnl}}
\makeatother
\RestyleAlgo{ruled}

\begin{document}

\title{Partial End-to-end Reinforcement Learning for Robustness Against Modelling Error in Autonomous Racing}

\author{Andrew Murdoch, Hendrik Willem Jordaan, Johannes Cornelius Schoeman
\thanks{
This work was supported by Stellenbosch University (Corresponding author: Andrew Murdoch, email: 20734751@sun.ac.za).
\hspace{0.1cm} The authors are with the Electrical and Electronic Engineering Department, Stellenbosch University, South Africa.
}}



\maketitle


\begin{abstract}

In this paper, we address the issue of increasing the performance of reinforcement learning (RL) solutions 
for autonomous racing cars when navigating under conditions where practical vehicle modelling errors (commonly known as \emph{model mismatches}) are present. 
To address this challenge, we propose a partial end-to-end algorithm that decouples the planning and control tasks. 
Within this framework, an RL agent generates a trajectory comprising a path and velocity, 
which is subsequently tracked using a pure pursuit steering controller and a proportional velocity controller, respectively. 
In contrast, many current learning-based (i.e., reinforcement and imitation learning) algorithms utilise an end-to-end approach
whereby a deep neural network directly maps from sensor data to control commands.
By leveraging the robustness of a classical controller, 
our partial end-to-end driving algorithm exhibits better robustness towards model mismatches than standard end-to-end algorithms.

\end{abstract}




\section{Introduction}
\IEEEPARstart{W}HILE autonomous vehicle research predominantly focuses on handling routine public road driving scenarios such as lane changing \cite{Moridpour2010}, 
ensuring complete safety demands the development of driving algorithms that can control vehicles at their handling limits. 
Autonomous racing competitions, such as F1tenth \cite{Babu2020} and Indy Autonomous Challenge \cite{Wischnewski2022}, 
are emerging as testing grounds for these algorithms \cite{Weiss2020}. 
Within the realm of autonomous racing, algorithms are tasked with processing LiDAR and odometry data to generate precise steering and throttle 
commands to navigate a vehicle around a circuit safely and in the fastest possible time \cite{Babu2020}.

Driving algorithms rely on simulated vehicle dynamics to generate feasible plans, as well as output control commands.
However, due to the difficulty involved in accurately modelling the physical vehicle, the simulated vehicle dynamics are often misaligned with the real-world vehicle dynamics \cite{Zhou2020}. 
This phenomenon, known as \emph{model mismatch}, causes the driving algorithm to underperform when tasked with controlling a real vehicle.
This disparity in performance is commonly referred to as the \emph{simulation-to-reality} gap.
In worst-case scenarios, the disparity between simulated and real vehicle dynamics can lead to the driving algorithm making a catastrophic error \cite{Ghignone2022}.
It is therefore imperative to ensure that autonomous racing algorithms exhibit robustness towards these modeling errors.

Classical approaches to solving autonomous racing generate robustness towards model mismatches by decoupling the driving task into the well-studied perception, planning, and control framework.
This approach has been used to develop racing algorithms capable of controlling full-scale racing cars \cite{Valls2018, alvarez2022, Nekkah2020} at speeds greater than 200 km/h.
Their success on physical vehicles is due to their ability to handle uncertainty in the vehicle model through a feedback controller.
For instance, \cite{Coulter_1992, Becker2022, Kritayakirana2010, Kritayakirana2012} and \cite{Hoffmann2007} present steering controllers that reliably minimise the vehicle's the cross track error, even in the presence of model mismatches.

Whilst classical methods handle uncertainty in the vehicle model well, they typically use optimisation algorithms for planning \cite{Heilmeier2020, Kelly2010, Liniger2015}.
These optimisation methods are neither computationally tractable \cite{Tatulea-Codrean2020}, nor robust towards environment uncertainty.
As such, they do not generalise well to more complex tasks such as racing with multiple vehicles \cite{Liniger2014}.
The limitations of classical approaches have motivated research into RL systems for autonomous racing.

RL approaches have achieved excellent results in scenarios that are considered challenging for classical approaches, 
such as racing with low computational power \cite{Evans2021a, Tatulea-Codrean2020} and racing against multiple vehicles \cite{Song2021, Wurman2022}.
RL algorithms are commonly implemented within an end-to-end architecture whereby a deep neural network (DNN) is trained to map sensor data directly to actuator commands \cite{Betz2021}.
The lack of any feedback control makes the end-to-end architecture sensitive to perturbations in the vehicle model, especially after training \cite{Fuchs2021}.
As such, results are so far limited to video games or small-scale F1tenth vehicles \cite{Fuchs2021, Ivanov2020, Chisari2021}.

With the racing domain in mind, this paper seeks to address the challenge of creating robust reinforcement learning autonomous racing algorithms capable 
of handling disparities between the vehicle dynamics during and after training.

\subsection{Related Work}

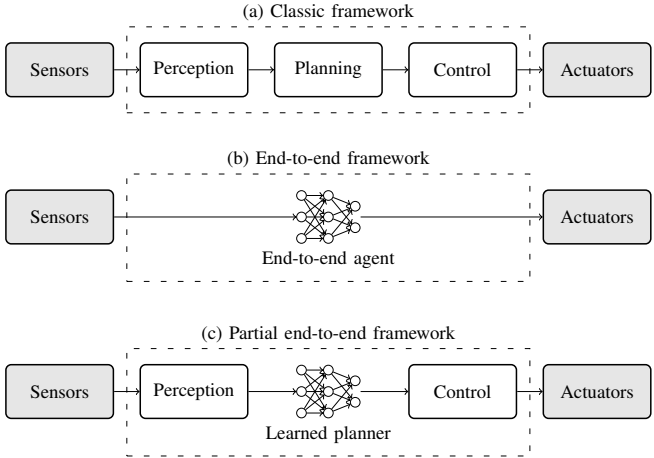
\begin{figure}[htb!]
    \centering
   \begin{subfigure}[htb!]{\columnwidth}
        \centering
        \def\svgwidth{\columnwidth}
        \resizebox{\columnwidth}{!}{
\tikzstyle{b}=[rectangle, draw=black, thick, minimum width = 2cm, minimum height = 1cm, align=center, rounded corners=.1cm, font=\normalsize]

\begin{tikzpicture}
 
    \node (a) at (0,0)  [b, fill=gray!20] {Sensors};
    \node (b) at (2.5,0)  [b] {Perception};
    \node (c) at (5,0)  [b] {Planning};
    \node (d) at (7.5,0)  [b] {Control};
    \node (e) at (10,0)  [b, fill=gray!20] {Actuators};
    
    \draw[loosely dashed] ($(b)+(-1.25,-0.8)$) -| ($(d)+(1.25,0.8)$) -|  ($(b)+(-1.25,-0.8)$) node [pos=0.25,above] {\normalsize (a) Classic framework};
    
    \draw[->] (a) -- (b);
    \draw[->] (b) -- (c);
    \draw[->] (c) -- (d);
    \draw[->] (d) -- (e);

\end{tikzpicture}}
        \label{fig:pete}
    \end{subfigure}
    \hfill
    \begin{subfigure}[htb!]{\columnwidth}
        \centering
        \def\svgwidth{\columnwidth}
        \resizebox{\columnwidth}{!}{
\tikzstyle{b} = [rectangle, draw=black, fill=white,thick, minimum width = 2cm, minimum height = 1cm, align=center, rounded corners=.1cm, font=\normalsize]
\tikzstyle{c} = [shape aspect=1, circle, minimum width=1.8mm, draw=black, inner sep=0pt]

\begin{tikzpicture}
 
    \node (a) at (0,0)      [b, fill=gray!20] {Sensors};
    \node (b) at (2.5,0)    [] {};
    \node (c) at (5,0)      [] {};
    \node (d) at (7.5,0)    [] {};
    \node (e) at (10,0)     [b, fill=gray!20] {Actuators};

    \node (nnl1r1) at ($(c)+(-0.5,0.4)$) [c] {};
    \node (nnl1r2) at ($(c)+(-0.5,0)$) [c] {};
    \node (nnl1r3) at ($(c)+(-0.5,-0.4)$) [c] {};
    
    \node (nnl2r1) at ($(c)+(0,0.4)$) [c] {};
    \node (nnl2r2) at ($(c)+(0,0)$) [c] {};
    \node (nnl2r3) at ($(c)+(0,-0.4)$) [c] {};
    \node (nnl3r1) at ($(c)+(0.5,0.2)$) [c] {};
    \node (nnl3r2) at ($(c)+(0.5,-0.2)$) [c] {};

    \node (nnl2r3Label) at ($(c)+(0,-1.2)$) [font=\normalsize, label={\normalsize End-to-end agent}] {};
    
    \draw[loosely dashed] ($(b)+(-1.25,-1.2)$) -| ($(d)+(1.25,0.8)$) -|  ($(b)+(-1.25,-1.2)$) node [pos=0.25,above] {\normalsize (b) End-to-end framework};
    
    \draw[->] (a) -- (nnl1r2);
    \draw[->] ($(c)+(0.6,0)$) -- (e);
    
    \draw[->] (nnl1r1) -- (nnl2r1);
    \draw[->] (nnl1r1) -- (nnl2r2);
    \draw[->] (nnl1r1) -- (nnl2r3);

    \draw[->] (nnl1r2) -- (nnl2r1);
    \draw[->] (nnl1r2) -- (nnl2r2);
    \draw[->] (nnl1r2) -- (nnl2r3);

    \draw[->] (nnl1r3) -- (nnl2r1);
    \draw[->] (nnl1r3) -- (nnl2r2);
    \draw[->] (nnl1r3) -- (nnl2r3);

    \draw[->] (nnl2r1) -- (nnl3r1);
    \draw[->] (nnl2r1) -- (nnl3r2);
    
    \draw[->] (nnl2r2) -- (nnl3r1);
    \draw[->] (nnl2r2) -- (nnl3r2);

    \draw[->] (nnl2r3) -- (nnl3r1);
    \draw[->] (nnl2r3) -- (nnl3r2);
    
\end{tikzpicture}}
        \label{fig:pete_learned_trajectory_planning}
    \end{subfigure}
    \hfill
    \begin{subfigure}[htb!]{\columnwidth}
        \centering
        \def\svgwidth{\columnwidth}
        \resizebox{\columnwidth}{!}{
\tikzstyle{b} = [rectangle, draw=black, fill=white,thick, minimum width = 2cm, minimum height = 1cm, align=center, rounded corners=.1cm, font=\normalsize]
\tikzstyle{c} = [shape aspect=1, circle, minimum width=1.8mm, draw=black, inner sep=0pt]

\begin{tikzpicture}
 
    \node (a) at (0,0)      [b, fill=gray!20] {Sensors};
    \node (left) at (2.5,0)    [b] {Perception};
    \node(center) at (5,0)  [] {};
    \node (c) at ($(center)+(0,0)$) [] {};
    \node (d) at ($(center)+(2.5,0)$) [b] {Control};
    \node (right) at (7.5,0)    [] {};
    \node (e) at (10,0)     [b, fill=gray!20] {Actuators};

    \node (nnl1r1) at ($(c)+(-0.5,0.4)$) [c] {};
    \node (nnl1r2) at ($(c)+(-0.5,0)$) [c] {};
    \node (nnl1r3) at ($(c)+(-0.5,-0.4)$) [c] {};
    
    \node (nnl2r1) at ($(c)+(0,0.4)$) [c] {};
    \node (nnl2r2) at ($(c)+(0,0)$) [c] {};
    \node (nnl2r3) at ($(c)+(0,-0.4)$) [c] {};
    \node (nnl3r1) at ($(c)+(0.5,0.2)$) [c] {};
    \node (nnl3r2) at ($(c)+(0.5,-0.2)$) [c] {};

    \node (nnl2r3Label) at ($(c)+(0,-1.2)$) [label={[align=center]\normalsize Learned planner}] {};
    
    \draw[loosely dashed] ($(left)+(-1.25,-1.2)$) -| ($(right)+(1.25,0.8)$) -|  ($(left)+(-1.25,-1.2)$) node [pos=0.25,above] {\normalsize (c) Partial end-to-end framework};
    
    \draw[->] (a) -- (left);
    \draw[->] (left) -- (nnl1r2);
    \draw[->] ($(c)+(0.6,0)$) -- (d);
    \draw[->] (d) -- (e);
    
    \draw[->] (nnl1r1) -- (nnl2r1);
    \draw[->] (nnl1r1) -- (nnl2r2);
    \draw[->] (nnl1r1) -- (nnl2r3);

    \draw[->] (nnl1r2) -- (nnl2r1);
    \draw[->] (nnl1r2) -- (nnl2r2);
    \draw[->] (nnl1r2) -- (nnl2r3);

    \draw[->] (nnl1r3) -- (nnl2r1);
    \draw[->] (nnl1r3) -- (nnl2r2);
    \draw[->] (nnl1r3) -- (nnl2r3);

    \draw[->] (nnl2r1) -- (nnl3r1);
    \draw[->] (nnl2r1) -- (nnl3r2);
    
    \draw[->] (nnl2r2) -- (nnl3r1);
    \draw[->] (nnl2r2) -- (nnl3r2);

    \draw[->] (nnl2r3) -- (nnl3r1);
    \draw[->] (nnl2r3) -- (nnl3r2);
    
\end{tikzpicture}}
        \label{fig:pete_learned_trajectory_planning}
    \end{subfigure}
    \caption{The common architectures utilised by autonomous driving algorithms. (a) The classic framework decouples perception, planning and control. 
    (b) End-to-end approaches utilise a DNN to perform the entire driving task, 
    and (c) approaches utilising the partial end-to-end framework use a DNN within the structure of the classic framework.}
    \label{fig:frameworkComparison}
\end{figure}

Prior work in the domain of learning algorithms applied to autonomous racing can be grouped into end-to-end and partial end-to-end frameworks.
Whereas end-to-end algorithms seek to replace the entire driving algorithm with a single DNN, partial end-to-end algorithms utilise the decoupled structure of classical algorithms, 
while combining or replacing modules with a DNN.
Figure \ref{fig:frameworkComparison} illustrates the differences between the common driving algorithm architectures.

\emph{\textbf{1) End-to-end Approaches:}}
Several techniques are employed to combat the effects that model mismatches have on the performance of end-to-end RL agents.
These are zero-shot transfer, one-shot transfer, domain randomisation, and policy smoothing.

In zero-shot transfer, the agent is tasked with controlling the physical vehicle without verifying that the simulation used during training matches reality.
It represents the simplest way to handle the simulation-to-reality gap \cite{brunnbauer2021}.
A more robust technique for deploying agents onto physical vehicles is one-shot transfer, 
whereby the simulation used to train the agent is verified to match reality via system identification techniques \cite{Ivanov2020}.
However, even if accurate system identification is performed, vehicle dynamics change over time.
For example, there are large disparities between road surface friction coefficients for dry and wet surfaces \cite{Zhao2017}.
Therefore, accurate system identification alone is insufficient. 
It is imperative to ensure that autonomous racing algorithms exhibit robustness towards the modelling errors that do occur.

Domain randomisation is performed by perturbing the vehicle model used by the simulator during training \cite{Josh2017}.
In the context of autonomous racing, two domain randomisation techniques are to sample the road friction coefficient from a normal distribution \cite{Ghignone2022}, 
or adding Gaussian noise to the lateral force experienced by the tires \cite{Chisari2021}.
The agents from these studies experienced a decrease in performance and safety when transferred onto a physical vehicle.
This is because the optimal policy for autonomous racing is highly sensitive to the vehicle model. 
As such, learning a policy that performs well across a range of vehicle models is challenging.

Policy output smoothing is utilised by \cite{Mysore2020, hsu2022} and \cite{brunnbauer2021} to ensure that their agents do not select extreme and jerky control actions.
Extreme steering and acceleration control actions cause the vehicle to operate closer to the friction limits of the tires than with smooth control actions, 
and can lead to uncontrollable and dangerous behaviour. 
This issue becomes more pertinent when considering scenarios in which model mismatches are present \cite{Chisari2021}.
Despite these efforts, end-to-end techniques still experience a decrease in performance and safety in the presence of model mismatches.
As such, addressing performance in model mismatch scenarios may require the incorporation of ideas from classical approaches, as in partial end-to-end systems.

\emph{\textbf{2) Partial End-to-end Approaches:}}
Two popular partial end-to-end design philosophies are to use a DNN to perform the task of either the planner \cite{Capo2020, Weiss2020, Mahmoud2020}, 
or that of the controller \cite{Evans2021b, Ghignone2022}.

The study by Ghignone et al. \cite{Ghignone2022} presents a learning-based system with excellent robustness to model mismatch by training an RL agent to learn the task of the controller.
This controller follows a trajectory generated by a model predictive contouring controller (MPCC).
However, the advantages of learning-based systems are their ability to learn complex behavior, and relegating the RL agent to the task of path following largely negates this benefit. 
Furthermore, classical controllers can already reliably achieve path following.

Several partial end-to-end approaches \cite{Capo2020, Weiss2020, Mahmoud2020} have utilized an algorithm structure whereby an RL agent is used for planning in conjunction with a classic controller for path tracking. 
These systems benefit from the agent’s heuristic nature in constructing the plan, while also leveraging the reliability of classical controllers to follow the path.
They have consistently outperformed end-to-end systems by a significant margin in simulation studies. 
However, their results have not been validated under conditions where model mismatches are present.

\subsection{Summary of contributions}
In this paper, we present a partial end-to-end algorithm that utilises an RL agent to generate a trajectory, which is then tracked using classic controllers.
In contrast to the studies by \cite{Capo2020, Weiss2020, Mahmoud2020},
we evaluate the performance of the partial end-to-end algorithm under conditions where model mismatches in the road surface friction coefficient, the vehicle mass 
and tire parameters are present.
Our results show that a learning algorithm that decouples planning and control experiences fewer crashes than end-to-end techniques when racing in model mismatch settings.
A further benefit of our approach over end-to-end methods is that the trajectory selected by the agent is constrained to the track limits.
This improves training time and performance, as well as improves the performance of the agent over end-to-end approaches when racing on more complex tracks.

\subsection{Structure of paper}
The remainder of this paper is structured as follows:
we discuss the implementation of an end-to-end algorithm
which is used to compare against the performance of our proposed solution. 
This baseline was chosen as similar end-to-end approaches are commonly used to solve racing problems \cite{Fuchs2021, Song2021}.
This is followed by a description of our proposed partial end-to-end algorithm. 
The race setting that we consider is then introduced, followed
by a discussion of how RL is applied to train the agents.
The fully and partial end-to-end algorithms' performances are then compared under conditions
with and without model mismatch present.


\section{End-to-end Algorithm}\label{ete}

\begin{figure}[htb!]
    \centering
    \tikzstyle{a}=[rectangle, draw=black, fill=gray!10,thick, minimum width = 2cm, minimum height = 2cm, align=center, rounded corners=.1cm, font=\small]
\tikzstyle{b}=[rectangle, draw=black, fill=gray!10,thick, minimum width = 1cm, minimum height = 0.8cm,, align=center, rounded corners=.1cm, font=\small]

\begin{tikzpicture}
 
    \node (a) at (5,5)  [a] {End-to-end\\agent};
    \node (b) at (8.5,5.5)  [b] {Velocity\\constraint};
    \node (c) at (8.5,3)  [b] {Environment};

    
    \draw[<-] (b) -- (b-|a.east);
    
    \draw[->] (b) -| ($(c)+(2,-0.2)$) -- ($(c.east)+(0,-0.2)$);
    \draw[->] ($(a.east)+(0,-0.5)$) -| ($(c)+(1.5,0.2)$) -- ($(c.east)+(0,0.2)$);
    \draw[->] (c) -| ($(a)+(-1.5,0)$) -- (a);

    \node (l1) at ($(a.east)+(0,0.5)$) [above right] {$a_{\text{long},d}$};
    \node (l3) at ($(b.east)+(0,0)$) [above right, fill=white] {$a_{\text{long}}$};
    \node (l2) at ($(a.east)+(0,-0.5)$) [above right] {$\delta_{d}$};
    \node (l4) at (c.west) [above left] {\small Observation};

   

    
\end{tikzpicture}
    \caption{The end-to-end algorithm architecture consists of an RL agent which outputs control commands, as well as a velocity constraint.}
    \label{fig:ete_architecture}
\end{figure}
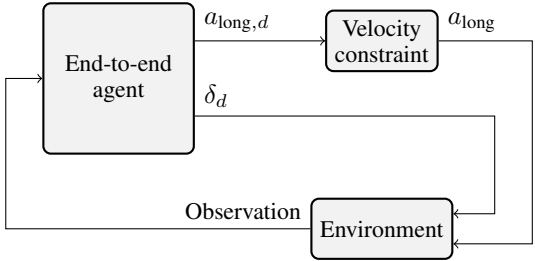

The end-to-end autonomous racing algorithm is composed of an RL agent and velocity constraint (similar to \cite{hsu2022}), as shown in Figure \ref{fig:ete_architecture}.
The RL agent is a DNN that maps directly from an observation to control commands.
These control commands are the desired longitudinal acceleration ($a_{\text{long},d}$) and the desired steering angle ($\delta_d$).
While the steering angle is passed directly to the simulator, the longitudinal action is modified by a velocity constraint component described by 
\begin{equation}
    a_{\text{long}} \leftarrow
    \begin{cases}
    0                   &   \text{for } v \geq v_{\text{max}}, \\
    0                   &   \text{for } v \leq v_{\text{min}}, \\
    a_{\text{long},d}   &   \text{otherwise},
    \end{cases}
\label{eq:speed_limit}
\end{equation}
where $v_{\text{max}}$ and $v_{\text{min}}$ are the imposed maximum and minimum allowable velocities listed in Table \ref{tab:constraint_parameters}, respectively.
This ensures that the velocity of the vehicle remains within safe bounds.

\section{Partial End-to-end Algorithm}

\begin{figure}[htb!]
    \centering
    \tikzstyle{a}=[rectangle, draw=black, fill=gray!10, thick, minimum width = 1cm, minimum height = 2.3cm, align=center, rounded corners=.1cm, font=\small]
\tikzstyle{b}=[rectangle, draw=black, fill=gray!10, thick, minimum width = 1cm, minimum height = 0.8cm, align=center, rounded corners=.1cm, font=\small]

\begin{tikzpicture}
 
    \node (a) at (5,5)  [a] {Planner\\agent};
    \node (b) at (10,5.7)  [b] {Velocity\\constraint};
    \node (c) at (8.65,2.7)  [b] {Environment};

    \node (d) at (7.3,5.7)  [b] {Velocity\\control};
    \node (e) at (7.3,4.3)  [b] {Steering\\control};
    
    
    \draw[<-] (e) -- (e-|a.east);
    \draw[<-] (d) -- (d-|a.east);
    \draw[<-] (b) -- (b-|d.east);
    \draw[->] (b) -| ($(c)+(2.7,-0.2)$) -- ($(c.east)+(0,-0.2)$);
    \draw[->] (c) -| ($(a)+(-1,0)$) -- (a);
    \draw[->] (e.east) -| ($(c)+(2,0.2)$) -- ($(c.east)+(0,0.2)$);

    \node (l1) at ($(a.east)+(0,0.7)$) [above right] {\small $v_d$};
    \node (l1) at ($(a.east)+(0,-0.7)$) [above right] {\small Path};
    \node (l3) at (e.east) [above right] {\small $\delta_{d}$};
    \node (l4) at (d.east) [above right] {\small $a_{\text{long},d}$};
    \node (l5) at (c.west) [above left] {\small Observation};
    \node (l6) at (b.east) [above right, fill=white] {\small $a_{\text{long}}$};

\end{tikzpicture}
    \caption{The partial end-to-end racing algorithm, which comprises an RL planner agent, velocity and steering controllers, as well as a velocity constraint.}
    \label{fig:steer_vel_architecture}
\end{figure}
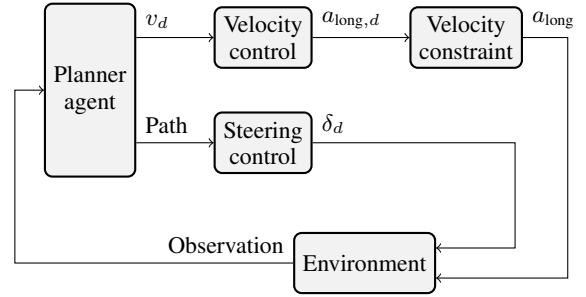

Our partial end-to-end algorithm comprises a planner RL agent, 
a steering and a velocity controller, as well as a velocity constraint, as depicted in Figure \ref{fig:steer_vel_architecture}.
The agent comprises a DNN that maps an observation to two parameters that are used to generate a trajectory at each time step.
One output of the RL agent is scaled to the range ($v_{\text{min}}, v_{\text{max}}$) and specifies the desired velocity (denoted as $v_{d}$).
The other output is used to generate a path.

\subsection{Path Generation}\label{sec:pathGeneration}

Similar to classical approaches \cite{Werling2010, Funke2017, Stahl2019}, 
the output of the agent is used to generate trajectories in a curvilinear coordinate system attached to the track centerline
known as the Frenet frame \cite{Werling2010}.
Within the Frenet frame, the distance along the horizontal axis corresponds to the distance along the centerline $s$. 
Furthermore, the vertical axis of the Frenet frame represents the perpendicular distance from the centerline $n$.
Navigating around the track is equivalent to traveling along the horizontal axis of the Frenet frame.
As a result, it is easier to create paths that avoid the track boundaries in Frenet coordinates compared to Cartesian coordinates, which proves valuable in preventing crashes.

\begin{figure}[htb!]
    \centering
    \def\svgwidth{\columnwidth}
    \resizebox{\columnwidth}{!}{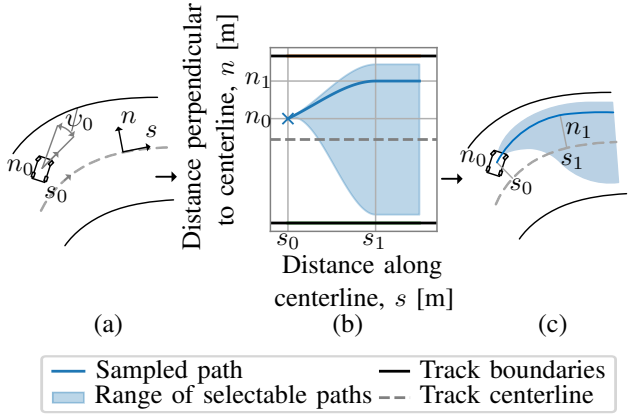}
    \caption[Generating the path in the Frenet frame]{An illustration of the process of generating the polynomial path in the Frenet frame. (a) The vehicle coordinates are converted into the Frenet frame, then (b) a path is constructed within the Frenet frame, after which (c) the path is converted into Cartesian coordinates.}
    \label{fig:polynomial_path_generation}
\end{figure}

Our specific approach to generating paths with the output of the RL planner agent is illustrated in Figure \ref{fig:polynomial_path_generation}, 
and involves solving a third-order polynomial function with constraints.
After converting the vehicle's coordinates and heading into the Frenet frame (denoted as $s_0$, $n_0$, and $\psi_0$, respectively), a third-order polynomial given by 
\begin{equation}
f(s) = As^3 + Bs^2 + Cs + D,
\end{equation}
is constructed. This polynomial is horizontally bounded by $s_0$ and $s_1$, where $s_1$ is chosen to be 2 meters ahead of $s_0$ along the centerline.
The following constraints are applied to the polynomial path;
\begin{enumerate}
    \item[(a)] The path passes through the vehicle's origin, satisfying $f(s_0) = n_0$.
    \item[(b)] At $s_0$, the path is parallel to the vehicle's heading, satisfying $f'(s_0) = \tan(\psi_0)$.
    \item[(c)] The perpendicular distance of the path from the centerline at $s_1$ is $n_1$, where $n_1$ is obtained by scaling the DNN output to the track width.
    This constraint is enforced by setting $f(s_1) = n_1$.
    \item[(d)] At $s_1$, the path is parallel to the centerline of the track, resulting in $f'(s_1) = 0$.
\end{enumerate}
Solving the polynomial using these constraints results in a path, 
which is then converted from Frenet to Cartesian coordinates for compatibility with the pure pursuit steering controller.

\subsection{Controllers}
The pure pursuit steering controller developed by Coulter \cite{Coulter_1992} is used to generate steering commands that track the generated path.
Meanwhile, a velocity controller generates desired acceleration commands, denoted $a_{\text{long},d}$, to ensure the vehicle maintains the velocity selected by the agent according to
\begin{equation}
    a_{\text{long},d} = 
    \begin{cases}
        k_v \frac{a_{\text{max}}}{v_{\text{max}}}(v_d - v) & \text{if } v_d \geq v\\
        k_v \frac{a_{\text{max}}}{v_{\text{min}}}(v_d - v) & \text{if } v_d < v,
    \end{cases}
\label{eq:vel_control}
\end{equation}
where $k_v$ is the gain, $a_{\text{max}}$ is the maximum acceleration listed in Table \ref{tab:constraint_parameters}, and $v$ is the vehicle's current velocity.
The velocity constraint described by Equation described by Equation \ref{eq:speed_limit} is then applied to the acceleration value given by the controller 
to ensure that the vehicle remains within safe operating limits.


\section{Reinforcement learning applied to train autonomous racing algorithms}

Partial and fully end-to-end agents are trained using a custom-built racing simulation environment.
Furthermore, the Twin delay deep deterministic policy gradient (TD3) algorithm by Fujimoto et al. \cite{Fujimoto2018} was used, 
due to its state-of-the-art performance on a variety of continuous control tasks.
After hyper-parameter tuning, the default TD3 hyperparameters given by Fujimoto et al. \cite{Fujimoto2018} were found to perform the best.
In addition to the simulator, actor and critic DNNs, as well as a reward signal are necessary to implement TD3.

\subsection{Simulation environment}
We developed a custom racing simulator inspired by F1tenth.
At each time step, this simulator accepts a control input (i.e., longitudinal acceleration and steering angle), 
and outputs an observation after transitioning the vehicle's state according to the single-track bicycle dynamics model by Althoff et al. \cite{Althoff2020}.
The observation includes the vehicle's pose ($x$ and $y$ coordinates, heading, and longitudinal velocity), 
as well as a LiDAR scan of $20$ equispaced beams with a field of view of $180\degree$.
Furthermore, the parameters and constraints of this bicycle model were identified for a standard F1tenth vehicle, and are listed in Appendix \ref{app:model}.

\subsection{Actor and Critic Networks}
Identical actor and critic DNNs are used for partial and fully end-to-end agents.
Each element of the observation vector is normalised to $(0,1)$ before being passed as input to the actor and critics.
Both actors and critics have an input layer comprising 400 ReLU-activated neurons, as well as a hidden layer of 300 ReLU-activated neurons. 
The output layer of the actor is activated with a hyperbolic tangent, whereas the output layers of the critics are activated linearly.
Due to the hyperbolic tangent activation, the actor's output is normalised to the range -1 and 1.
For the end-to-end agent, the output of the actor is scaled to the ranges of the control actions, 
\begin{equation}
    a_{\text{long},d} \in (-a_{\text{max}}, a_{\text{max}}), \hspace{0.2cm} \text{and} \hspace{0.2cm} \delta_{d} \in (\underline{\delta}, \overline{\delta}), 
\end{equation}
as listed in Table \ref{tab:constraint_parameters}.
However, the output of the partial end-to-end agent's actor corresponding to the longitudinal action is scaled to $(v_{\text{min}}, v_{\text{max}})$. 
Additionally, the output corresponding to the lateral action is used to generate the path as described in Section \ref{sec:pathGeneration}.

\subsection{Reward function}
The reward signal is designed to closely approximate the objective of minimizing lap time for high reward discount rates.
Specifically, the agent is rewarded for the distance it travels along the centerline between the current and previous time step, as described by \cite{Fuchs2021}.
In addition, the agent receives a small penalty at every time step, as well as a large penalty if it collides with the track boundary. 
The resulting piece-wise reward signal is expressed as
\begin{dmath}
r(s_t,a_t) = 
\begin{cases}
r_{\text{collision}} & \mbox{if collision occurred} \\
r_{\text{dist}}(D_{t} - D_{t-1}) + r_{\text{time}} & \mbox{otherwise,} \\
\end{cases}
\label{eq:reward_signal}
\end{dmath}
where $r_{\text{collision}}$, $r_{\text{dist}}$, and $r_{t}$ represent the penalty for collisions, the reward for distance traveled, and the penalty for each time step, respectively.
Additionally, $D_t$ and $D_{t-1}$ are the distance travelled along the centerline at the current and previous time steps.
Notably, this reward signal is similar to those used in numerous prior works \cite{Song2021, Ivanov2020, Perot2017, Jaritz2018, brunnbauer2021, Evans2021b}.



\section{Experiments and Results}

The race setting considered in our experiments is similar to the time-trial stage of the F1tenth competition.
That is, a single vehicle competes to complete one lap of one of the racetracks shown in Table \ref{tab:tracks}.
These racetracks include one simple track (Porto), as well as two more complicated and longer tracks which are modelled after real racing circuits (Barcelona-Catalunya and Monaco).
The start and finish line coincides with the starting position of the vehicle, which is chosen randomly along the length of the track at every episode. 
If the vehicle makes contact with the track boundary, which is detectable by the LiDAR scanner, the lap is considered failed, and the lap time is not counted.

\newcolumntype{R}{>{\raggedleft\arraybackslash}p{1.5cm}}
\newcolumntype{C}{>{\centering\arraybackslash}p{1.8cm}}

\begin{table}[htb!]
\centering
\renewcommand{\arraystretch}{1.7}
\small
\begin{tabular}{RCCC} 
    \hline
    \textbf{Metric} & \textbf{Porto} & \textbf{Barcelona-Catalunya} & \textbf{Monaco}\\ 
    \hline
    Length (m) & 30.7 & 236.8 & 178.3 \\
    Image &  \raisebox{-0.5\height}{\includegraphics[width=1.5cm]{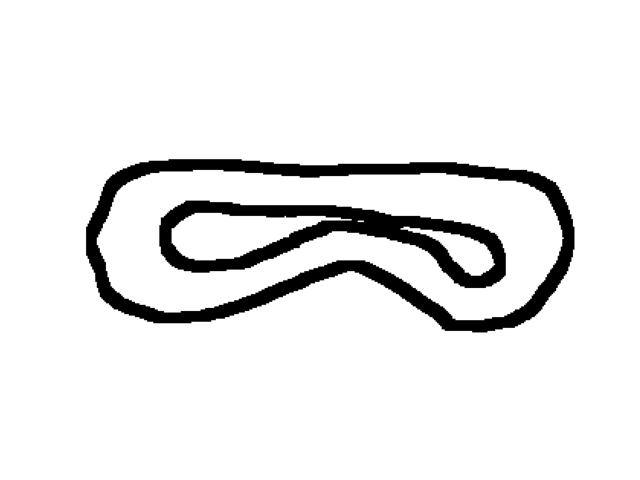}} &  \raisebox{-0.5\height}{\includegraphics[width=2.2cm]{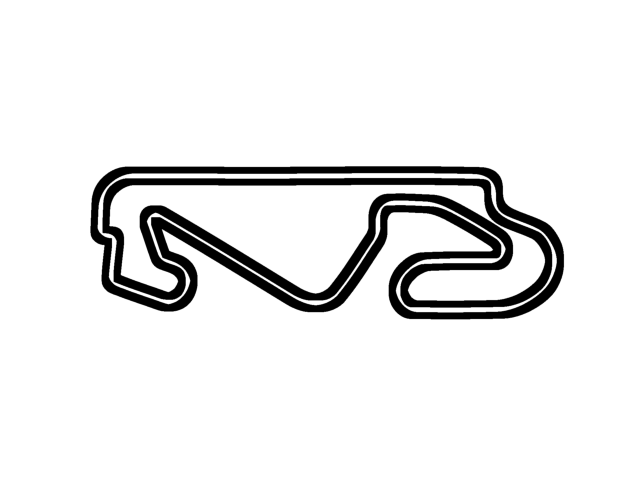}} &  \raisebox{-0.5\height}{\includegraphics[width=1.8cm]{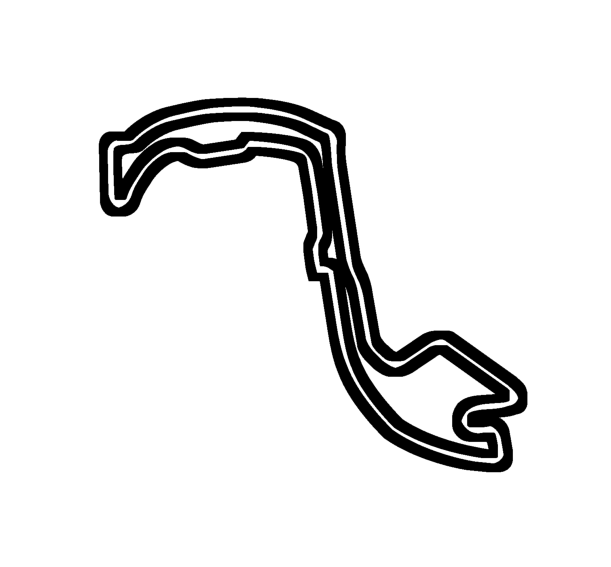}} \\
    \hline
\end{tabular}
\caption{Lengths and image of the the three tracks used to train and evaluate agents.}
\label{tab:tracks}
\end{table}

We assess the performance of fully and partial end-to-end algorithms during training, as well as under evaluation conditions during which model mismatch is not present.
This is followed by an investigation into the performance of these algorithms when racing under simulated model mismatch conditions.

\subsection{Training performance}

Agents were trained on a single track for a fixed number of MDP time steps. 
To introduce variation, the initial position of the vehicle along the track was randomized at each episode. 
The parameters of the vehicle dynamics model remained constant across episodes, ensuring there was no model mismatch during training. 
Additionally, no sensor noise or observation error was present in the simulation.

\begin{figure}[h]
    \centering
    \def\svgwidth{\columnwidth}
    \resizebox{\columnwidth}{!}{\input{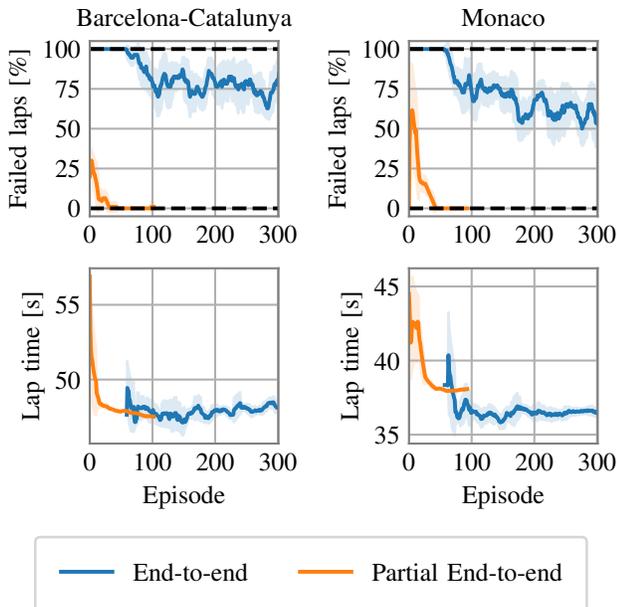}}
    \caption{Percentage failed laps and average lap time of 10 partial and 10 fully end-to-end agents learning to race on the Barcelona-Catalunya track (left), as well as Monaco (right).}
    \label{fig:BarcelonaTraining}
\end{figure}

The percentage of failed laps and average lap time for 10 partial and 10 fully end-to-end agents learning to race on the Barcelona-Catalunya and Monaco tracks are shown in Figure \ref{fig:BarcelonaTraining}.
A trend noticed across both tracks is that end-to-end agents continue to crash throughout training, whereas partial end-to-end agents reach a $0\%$ crash rate early in training.
For instance, end-to-end agents learning to race on the Barcelona-Catalunya track successfully completed only $47.9\%$ of their training episodes without crashing, whereas partial end-to-end agents successfully completed $90.3\%$ of their training episodes.
Furthermore, the training time needed for partial end-to-end agents to reach competitive performance levels is much shorter than end-to-end agents.
Interestingly, both partial and fully end-to-end agents achieve similar lap times for the laps that are successfully completed.

\subsection{Racing without model mismatch}

After training each agent, we assessed their performances under conditions
whereby noise is present in the observation, but no exploration noise is added to the agent's selected actions.
Each of the 10 partial and 10 fully end-to-end agents completed 100 laps on the same track as it was trained on under these evaluation conditions.
Table \ref{tab:evaluation} lists the percentage of successful laps and average lap time of the algorithms racing on each track.

\newcolumntype{R}{>{\centering\arraybackslash}p{1.3cm}}
\newcolumntype{C}{>{\centering\arraybackslash}p{1.35cm}}

\begin{table}[htb!]
\centering
\renewcommand{\arraystretch}{1.3}
\small
\begin{tabularx}{\columnwidth}{RCCCC} 
    \hline
    & \multicolumn{4}{c}{\textbf{Algorithm}} \\
    \cmidrule(lr){2-5}
    \textbf{Track} &  \multicolumn{2}{c}{\textbf{End-to-end}} &  \multicolumn{2}{c}{\textbf{Partial end-to-end}} \\
    \cmidrule(lr){2-3}  \cmidrule(lr){4-5}
    & Successful laps [$\%$] & Lap time [s] & Successful laps [$\%$] & Lap time [s] \\
    \hline
    Porto & 98.9 & 6.05 & \textbf{100.0} & \textbf{5.86} \\
    Barcelona-Catalunya & 56.3 & 47.39 & \textbf{99.9} & \textbf{47.12} \\
    Monaco & 59.2 & \textbf{35.63} & \textbf{100.0} & 37.91 \\
    \hline
\end{tabularx}
\caption{Performance of end-to-end and partial end-to-end agents racing on all three tracks under evaluation conditions. Bold values indicate the best performance.}
\label{tab:evaluation}
\end{table}

Table \ref{tab:evaluation} shows that end-to-end agents are competitive with partial end-to-end agents in terms of lap time and success rate when tasked with racing on a simple track such as Porto.
However, partial end-to-end agents offer a significant advantage in terms of success rate when the complexity of the track is increased.
In fact, our approach outperformed the end-to-end baseline in terms of the percentage of successful laps completed in every setting considered.

\begin{figure*}[htb!]
    \centering
    \def\svgwidth{\textwidth}
    \resizebox{\textwidth}{!}{\input{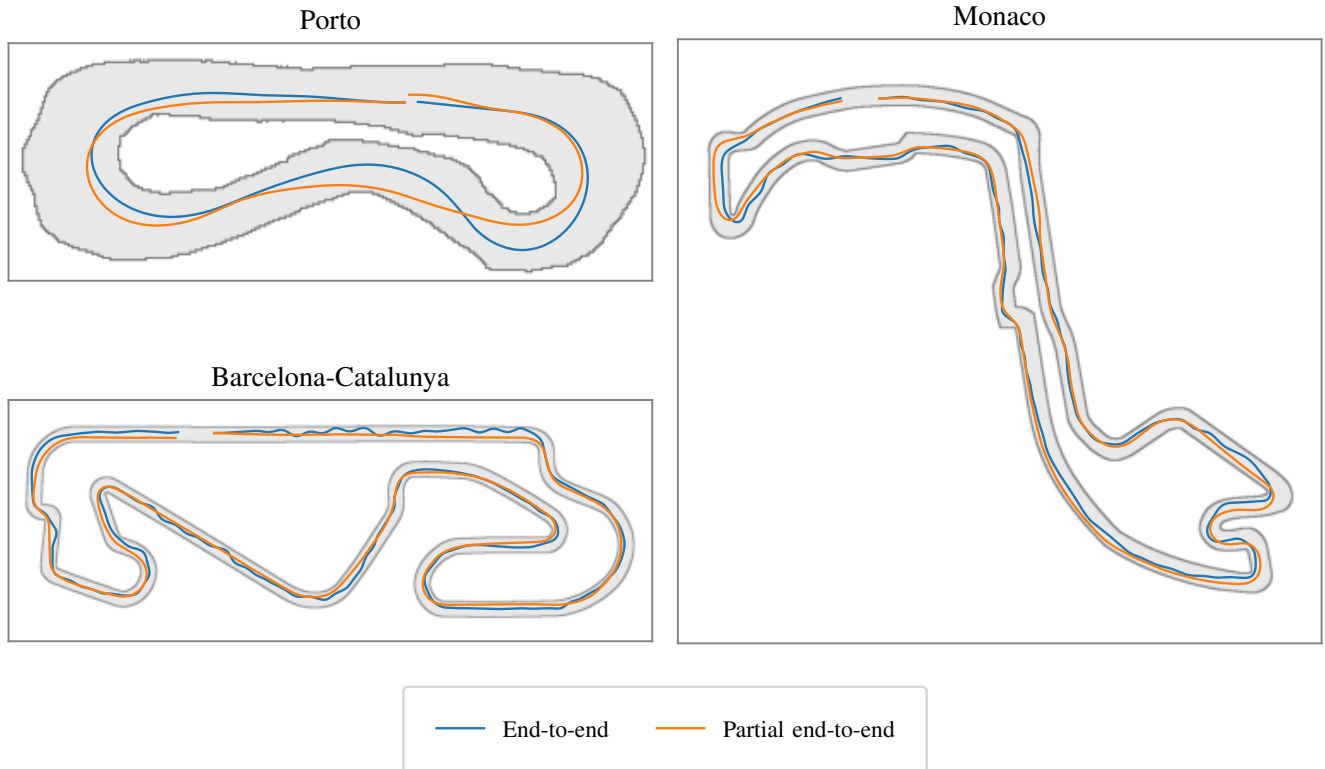}}
    \caption{Trajectories executed by a partial and a fully end-to-end agent racing on the Porto, Barcelona-Catalunya and Monaco circuits under ideal evaluation conditions 
    (i.e., no model mismatch is present).}
    \label{fig:trajectoryGrid}
\end{figure*}

In addition to comparing the success rates and average lap times quantitatively, the trained agents' driving behaviour is analysed qualitatively.
Figure \ref{fig:trajectoryGrid} shows trajectories executed by a partial and a fully end-to-end agent on all three tracks considered.
Whilst partial and fully end-to-end agents achieve similar performances when on the simple Porto track, the quality of trajectories executed by end-to-end agents decreases significantly when tasked with racing on complex tracks, as evidenced by the large degree of slaloming present in the agent's path.
However, the partial end-to-end agent demonstrated the ability to execute smooth trajectories, due to the constraints placed on the path.
Thus, embedded knowledge of the track geometry, as well as the ability to execute low-level control to follow the trajectories output by the RL planner agent is beneficial in ideal environments.

\subsection{Racing with model mismatch}

Our investigations encompass practical vehicle dynamics modeling errors stemming from three sources, namely;
\begin{enumerate}
\item vehicle mass and mass distribution,
\item tire cornering stiffness coefficient, and
\item road surface friction coefficient.
\end{enumerate}
These are types of model mismatches that are expected to be found in road-going cars. 
Thus, we consider the worst-case scenario in which the agent must operate the vehicle at its handling limits in the presence of significant model uncertainty.

To simulate model mismatch scenarios, we introduced variations in the vehicle model between the training and evaluation.
Agents are then evaluated according to the same procedure as without model mismatch.
This approach is similar to the ones taken by Fuchs et al. \cite{Fuchs2021} and Ghignone et al. \cite{Ghignone2022}.
The experiments were carried out using a single agent per algorithm architecture, with each chosen agent serving as a representative of the median performance within its respective framework.

Note that the notion of model mismatch refers to a discrepancy between the vehicle model used during training and evaluation.
Therefore, although online estimation of vehicle model parameters is possible, model mismatch persists unless the updated vehicle model can be utilized to either retrain the agent online, or the current agent can be replaced with another one that was trained with a set of vehicle parameters more similar to the real vehicle.
These approaches may be prohibitively expensive for scenarios where the driving task is complex.
Furthermore, while model mismatch is a phenomenon that takes place when deploying agents in the real world, our analysis is restricted to simulation so that the effect of each model mismatch type can be studied individually.

\emph{\textbf{1) Discrepancy in Road Surface Friction:}}
The performance of partial and fully end-to-end agents under conditions where the road surface friction coefficient value used to train the agents is erroneous was investigated.
Agents were evaluated with road surface friction values ranging from $0.5$ (representing a typical value for wet asphalt) to $1.04$ (corresponding to the nominal training value for F1tenth cars given in Table \ref{tab:vehicle_parameters}) on all three tracks. 
Importantly, the simulator assumes a spatially and temporally uniform road friction coefficient throughout a lap.
With this assumption, the worst-case scenario whereby the entire road surface has a decreased friction coefficient is considered.

\begin{figure}[htb!]
    \centering
    \def\svgwidth{\columnwidth}
    \resizebox{\columnwidth}{!}{\input{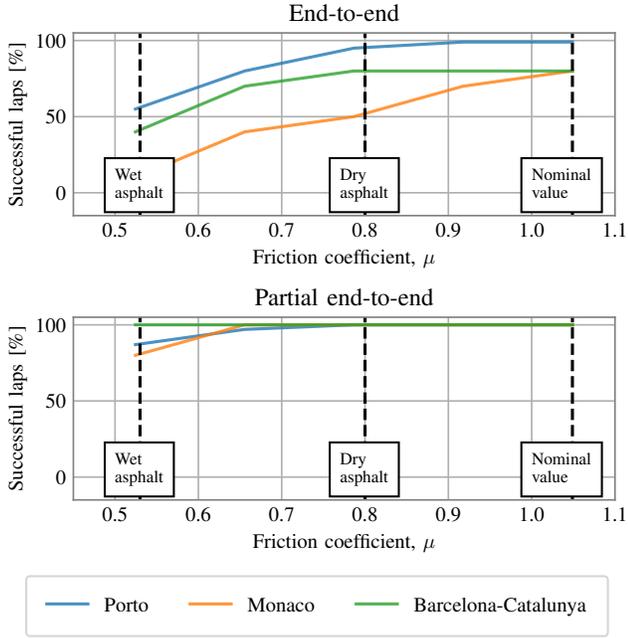}}
    \caption[Success rate of agents under evaluation conditions with mismatched road surface friction coefficient]{Success rates under evaluation conditions of agents racing with decreased road surface friction values. Results for the end-to-end agent racing on all three tracks are shown in the top subplot, while results for the corresponding results for partial end-to-end agents are shown on the bottom. The values of friction corresponding to the nominal training value, dry and wet asphalt are marked with a black dashed line.}
    \label{fig:mu}
\end{figure}

Figure \ref{fig:mu} illustrates the percentage of successful evaluation laps for both end-to-end and partial end-to-end agents when facing a mismatch in the road surface friction coefficient across all three tracks. 
Under nominal conditions, the end-to-end agents achieved success rates of $99\%$, $80\%$, and $79\%$ for Porto, Barcelona-Catalunya, and Monaco, respectively. However, when considering the worst-case scenario of racing on a wet asphalt surface, the success rates significantly decreased to $55\%$, $41\%$, and $11\%$ for these respective tracks.


On the other hand, the partial end-to-end agents successfully completed all their evaluation laps under nominal conditions. 
When a model mismatch in the road surface friction coefficient was introduced simulating a wet asphalt surface, the partial end-to-end agent was still able to complete all of its laps on Circuit de Barcelona-Catalunya.
However, its success rate decreased to  $87\%$ and $81\%$ for the Porto and Monaco tracks.
Furthermore, almost no crashes occur for any friction coefficient value above $0.7$ for any track.

\begin{figure}[htb!]
    \centering
    \def\svgwidth{\columnwidth}
    \resizebox{\columnwidth}{!}{\input{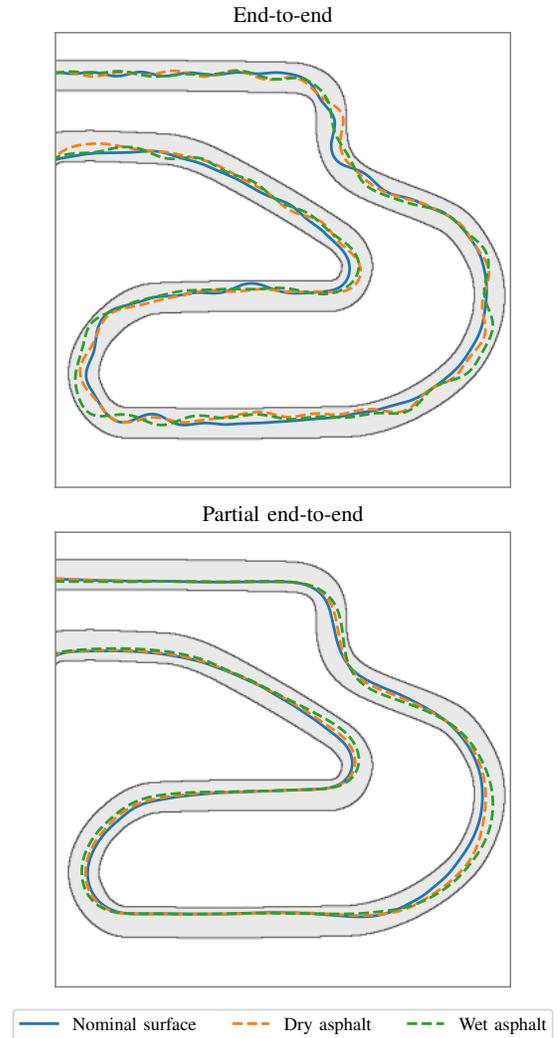}}
    \caption[Trajectories of agents racing with a decreased road surface friction coefficient]{Trajectories of agents racing on a section of Barcelona-Catalunya under nominal conditions as well as road surface friction values corresponding to dry asphalt and wet asphalt.}
     \label{fig:mu_lap}
\end{figure}

An example of trajectories executed by both partial and fully end-to-end agents racing on a section of Circuit de Barcelona-Catalunya is shown in Figure \ref{fig:mu_lap}.
The figure shows the trajectories executed by agents under nominal conditions, as well as conditions with decreased surface friction.
Note that the trajectories executed by the partial end-to-end agent remain similar when mismatches are introduced.
In fact, the most notable difference between trajectories executed by the partial end-to-end agent on the nominal and slippier surfaces is that the trajectories executed on the slippier surfaces exhibit reduced curvature, resulting in wider paths being followed by the agents.
In contrast, the trajectories by the end-to-end agent are always erratic.
Thus, the partial end-to-end agent outperformed the end-to-end agent on all road surface conditions considered.


\emph{\textbf{2) Uncertain Cornering Stiffness:}}
Another practical model mismatch that vehicles encounter is a discrepancy in the cornering stiffness terms ($C_{S,r}, C_{S,f}$ for the rear and front, respectively) which characterises their tires.
Tire construction and dimensions, the type and quality of the tread, and inflation pressure are significant factors when determining cornering stiffness \cite{Vorotovic2013}.

\begin{figure*}[htb!]
    \centering
    \def\svgwidth{\textwidth}
    \resizebox{\textwidth}{!}{\input{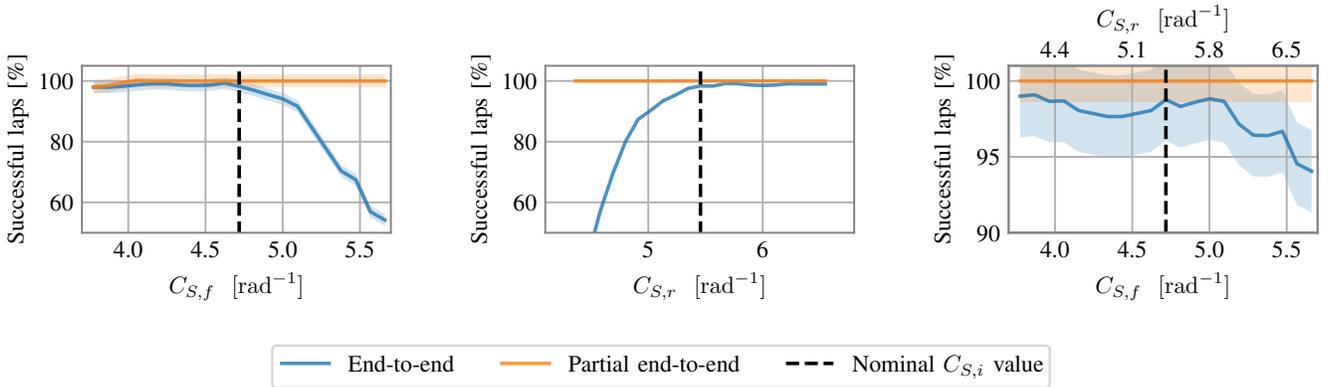}}
    \caption[Success rate of agents under evaluation conditions with mismatched tire cornering stiffness]{Success rates under evaluation conditions of agents with mismatched tire cornering stiffness. The left plot shows the effect of varying only the front tire stiffness. The effect of varying only rear tire stiffness, and both front and rear tire stiffness together are shown in the middle and right plots, respectively.}
    \label{fig:c_s}
\end{figure*}

An investigation was conducted whereby the performance of agents was analysed after simulating changes in cornering stiffness coefficient for (a) the front tires, (b) the rear tires, then (c) both the front and rear tires together.
During evaluation, the tire stiffnesses were varied by up to $\pm 20 \%$ from the nominal values of $4.72$ and $5.45$ $\text{rads}^{-1}$ for the front and rear tires, respectively.
Furthermore, we chose to perform this experiment on the Porto track, because the difference in performance between the partial and fully end-to-end agents under nominal 
(i.e., no model mismatch) conditions were minimal on this track.
The percentage of successful laps of agents racing with mismatched cornering stiffness coefficients on the Porto track are shown in Figure \ref{fig:c_s}.

As is evident from Figure \ref{fig:c_s}, the end-to-end agent is sensitive to an increase in the front cornering stiffness coefficient, while also being sensitive to a decrease in the rear cornering stiffness coefficient.
In addition, when both the front and rear cornering stiffness coefficients are altered simultaneously, the end-to-end agent tends to experience crashes. 
In contrast, the partial end-to-end agent demonstrates resilience to changes in either front or rear tires.
Although our partial end-to-end approach does experience failures when both the front and rear cornering stiffness coefficients are decreased together by $20\%$, the failure rate is low compared to end-to-end agents.


\begin{figure}[h]
    \centering
    \def\svgwidth{\columnwidth}
    \resizebox{\columnwidth}{!}{ \input{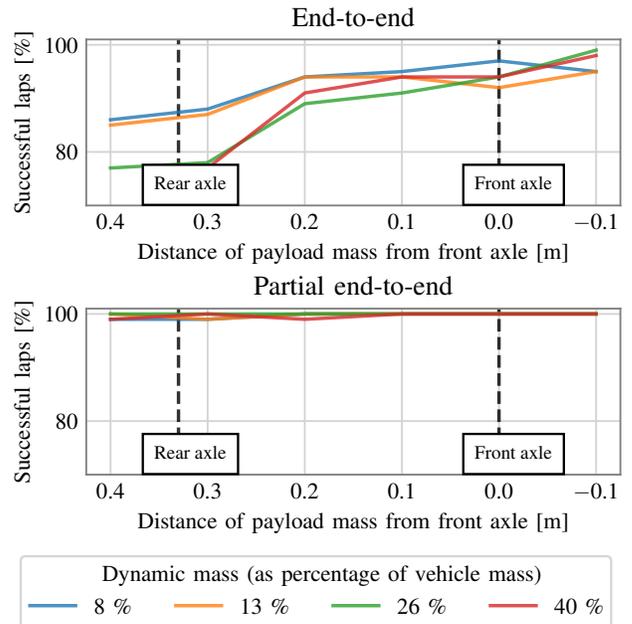}}
    \caption{The percentage of successful laps under evaluation conditions for agents with masses placed along the longitudinal axis of the vehicle. The front and rear axles are indicated by black dashed lines.}
    \label{fig:unknown_mass}
\end{figure}

\emph{\textbf{3) Adding a Dynamic Mass:}}
To investigate the effect of a model mismatch in vehicle mass, we simulated the addition of various masses along the longitudinal axis of the vehicle in between training and evaluation. 
These masses were 0.3 kg, 0.5 kg, 1 kg, and 1.5 kg (corresponding to $8 \%$,  $13 \%$, $26 \%$, and $40 \%$ of the vehicle's mass, respectively).
The percentage of successful laps achieved by the agents is plotted as a function of the position of the masses along the length of the vehicle in Figure \ref{fig:unknown_mass}.


Our findings indicate that the end-to-end agent exhibited sensitivity to an unaccounted-for mass, particularly when the mass was positioned toward the back of the vehicle. 
This observation is supported by the significant decrease in lap completion rate when the mass is placed closer to the rear axle. 
Interestingly, the end-to-end agent displayed some resilience towards a mass located at the front of the vehicle. 
In contrast, the partial end-to-end algorithm demonstrated a higher level of robustness against modeling errors stemming from unaccounted-for masses. 
Regardless of the position of the mass placement on the vehicle, the partial end-to-end agent successfully completed a large percentage of nearly all of its evaluation laps.


\section{Conclusion}

In this paper, we have presented a solution to the model mismatch problem faced by autonomous driving and racing algorithms.
Specifically, we aimed to improve the performance of current reinforcement learning (i.e., end-to-end) systems under conditions where model mismatches are present.
Our solution introduces a partial end-to-end reinforcement learning algorithm that incorporates ideas from classical algorithms that perform well under conditions whereby uncertainty is present in the vehicle model; namely, the decoupling of the planning and control tasks.

The partial end-to-end approach offers an additional advantage by enhancing performance irrespective of the presence of model mismatches. 
This improvement stems from the integration of track boundary information into the agent's action space.
By defining the path within the Frenet frame, our partial end-to-end approach successfully generates trajectories that consistently steer clear of track boundaries.
Furthermore, the use of classic feedback controllers offered better performance than an end-to-end approach to vehicle control in model mismatch conditions.
Notably, the partial end-to-end agents exhibited reduced crashes and even executed similar trajectories prior to and following the introduction of model mismatches.

The performance of partial end-to-end agents under nominal conditions, as well as conditions where model mismatches are present, 
highlights the potential of these algorithms in developing robust learning-based driving systems that can handle uncertainties in vehicle dynamics.
Furthermore, these results have implications for addressing vehicle safety in the broader road-going autonomous driving problem.





{\appendix
\subsection{Single Track Bicycle Dynamics Model}\label{app:model}

\begin{table}[h]
\centering
\begin{tabular}{rlll} 
    \hline
    \textbf{Symbol} & \textbf{Description} & \textbf{Unit} & \textbf{Value}\\ 
    \hline
    $m$ & Mass & kg & $3.74$ \\ 
    $I_z$ & Moment of inertia about z axis & kg & $0.04712$ \\
    $l_f$ & Distance from CoG to front axle & m & $0.1587$ \\
    $l_r$ & Distance from CoG to rear axle & m & $0.17145$ \\
    $h_{cg}$ & Height of CoG & m & $0.074$ \\
    $C_{S,f}$ & Cornering stiffness coefficient (front) & 1/rad & $4.718$ \\
    $C_{S,r}$ & Cornering stiffness coefficient (rear) & 1/rad & $5.4562$ \\
    $\mu$ & Road surface friction coefficient & - & $1.0489$ \\
    \hline
\end{tabular}
\caption[Vehicle model parameters for the single track model]{Vehicle model parameters for the single track model. The parameters were identified for a standard F1tenth vehicle.}
\label{tab:vehicle_parameters}
\end{table}

\begin{table}[h]
\centering
\small
\begin{tabular}{rlll} 
    \hline
    \textbf{Name} & \textbf{Symbol} & \textbf{Unit} & \textbf{Value}\\ 
    \hline
    Minimum steering angle & $\underline{\delta}$ & rad & $-0.4189$ \\
    Maximum steering angle & $\overline{\delta}$ & rad & $0.4189$ \\
    Minimum steering rate & $\underline{\dot{\delta}}$ & rad/s & $-3.2$ \\
    Maximum steering rate & $\overline{\dot{\delta}}$ & rad/s & $3.2$ \\
    Minimum velocity & $\underline{v}$ & m/s & $-5$ \\
    Maximum velocity & $\overline{v}$ & m/s & $20$\\
    Maximum acceleration & $a_{\text{max}}$ & m/s$^2$ & $9.51$ \\
    Switching velocity & $v_S$ & m/s & $7.319$\\
    Maximum allowable velocity & $v_{\text{max}}$ & m/s & 5 \\
    Minimum allowable velocity & $v_{\text{min}}$ & m/s & 3 \\
    \hline
\end{tabular}
\caption[Vehicle constraint parameters]{Constraint parameters of a standard F1tenth vehicle.}
\label{tab:constraint_parameters}
\end{table}

}


\bibliographystyle{IEEEtran}
\bibliography{library}

 

\begin{IEEEbiography}[{\includegraphics[width=1in,height=1.25in,clip,keepaspectratio]{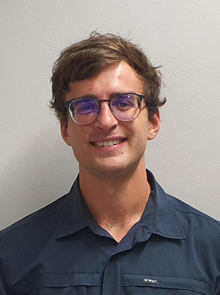}}]
{Andrew Murdoch}
received his bachelor's degree in mechatronic engineering from the Stellenbosch University in 2021, 
and continued studying to receive his master's in electronic engineering in 2023.
He forms part of the Electronic Systems Laboratory (ESL), where he will start with his Ph.D. in 2024.
His research interest is artificial intelligence applied to autonomous vehicles.
\end{IEEEbiography}

\begin{IEEEbiography}[{\includegraphics[width=1in,height=1.25in,clip,keepaspectratio]{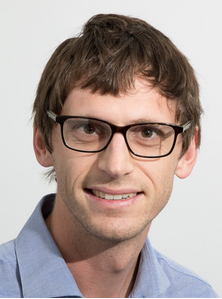}}]
{Willem Jordaan}
received his bachelor’s in Electrical and Electronic
Engineering with Computer Science and continued to receive his Ph.D. degree
in satellite control at Stellenbosch University. He currently acts as a senior
lecturer at Stellenbosch University and is involved in several research projects
regarding advanced control systems as applied to different autonomous
vehicles. His interests include robust and adaptive control systems applied
to practical vehicles.  He is a Senior Member of the IEEE.
\end{IEEEbiography}

\begin{IEEEbiography}[{\includegraphics[width=1in,height=1.25in,clip,keepaspectratio]{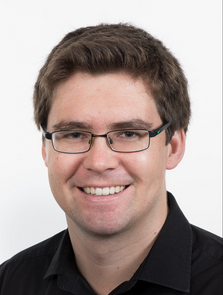}}]
    {JC Schoeman}
    is a Lecturer at the Department of Electrical and Electronic Engineering at Stellenbosch University. 
    He also obtained his PhD in 2021 from the same university, with a focus on "degenerate Gaussian factors for probabilistic inference". 
    His broad research interest lies at the intersection of robotics and machine learning, and more specifically in developing autonomous systems with optimal decision-making capabilities.
\end{IEEEbiography} 

\vfill

\end{document}